%% file: acl_latex.tex
\documentclass[11pt]{article}

\usepackage[preprint]{acl}

\usepackage{booktabs}
\usepackage{multirow} 
\usepackage{amssymb}
\usepackage{pifont}
\newcommand{\cmark}{\ding{51}}%

\usepackage{times}
\usepackage{latexsym}
\usepackage{subcaption}
\usepackage[T1]{fontenc}

\usepackage[utf8]{inputenc}

\usepackage{microtype}

\usepackage{inconsolata}

\usepackage{graphicx}

\usepackage{hyperref}

\usepackage[table]{xcolor}
\definecolor{suborange}{RGB}{255,240,220}
\definecolor{headerblue}{RGB}{230,240,255}
\definecolor{bestgreen}{RGB}{220,250,230}
\usepackage{color}
\definecolor{darkblue}{rgb}{0, 0, 0.5}
\hypersetup{colorlinks=true, citecolor=darkblue, linkcolor=darkblue, urlcolor=darkblue}
\definecolor{darkgreen}{RGB}{50,100,0}
\definecolor{darkred}{RGB}{200, 0, 0}
\definecolor{lightblue}{RGB}{220,235,250}
\usepackage[most,skins,theorems]{tcolorbox}
\tcbset{
  takeawaysbox/.style={
    title=Observation,
    colback=lightblue!80,
    colframe=black,
    fonttitle=\bfseries\small,
    coltitle=white,
    colbacktitle=black,
    enhanced,
    attach boxed title to top left={xshift=2.5mm,yshift=-2.5mm},
    boxed title style={rounded corners, size=small, colframe=black, colback=black},
    width=\linewidth,
    arc=3.5mm
  }
}
\usepackage{algorithm}
\usepackage{algpseudocode}

\usepackage{amsmath}
\usepackage{amssymb}

\usepackage[most]{tcolorbox}
\tcbuselibrary{skins,breakable,listings}
\definecolor{PromptBlue}{HTML}{2563EB} 
\definecolor{PromptBg}{HTML}{F5F9FF}
\newtcblisting{systemprompt}{
  enhanced,
  breakable,
  listing only,
  title={Reasoning Cue},
  fonttitle=\bfseries,
  colback=PromptBg,
  colframe=PromptBlue,
  colbacktitle=PromptBlue!7,
  coltitle=PromptBlue!70!black,
  boxrule=0.8pt,
  arc=3pt,
  left=6pt,right=6pt,top=6pt,bottom=6pt,
  attach boxed title to top left={yshift=-2mm, xshift=2mm},
  drop fuzzy shadow,
  listing options={
    basicstyle=\ttfamily\small,
    breaklines=true,
    columns=flexible,   
    tabsize=2,
    showstringspaces=false,
    xleftmargin=0pt,    
    aboveskip=0pt,
    belowskip=0pt,
    keepspaces=true     
  }
}

\usepackage{cleveref}
\title{Understanding and Mitigating Spurious Signal Amplification \\in Test-Time Reinforcement Learning for Math Reasoning}

\author{
 \textbf{Yongcan Yu\textsuperscript{1,2}\thanks{Work done during an internship at Meituan}},
 \textbf{Lingxiao He\textsuperscript{1,3}},
 \textbf{Jian Liang\textsuperscript{1,2}\thanks{Corresponding author}},
 \textbf{Kuangpu Guo\textsuperscript{1,4}},
\\
 \textbf{Meng Wang\textsuperscript{3}},
 \textbf{Qianlong Xie\textsuperscript{3}},
 \textbf{Xingxing Wang\textsuperscript{3}},
 \textbf{Ran He\textsuperscript{1,2}}
\\
\\
 \textsuperscript{1}NLPR \& MAIS, Institute of Automation, Chinese Academy of Sciences\\
 \textsuperscript{2}School of Artificial Intelligence, University of Chinese Academy of Sciences\\
 \textsuperscript{3}Meituan
 \textsuperscript{4}University of Science and Technology of China\\
\texttt{\{yuyongcan0223, liangjian92\}@gmail.com}
}

\begin{document}
\maketitle

\input{sections/abs}
\input{sections/intro}
\input{sections/observations}
\input{sections/method}

\input{sections/exp}

\input{sections/rel_work}
\input{sections/conclu}

\section*{Limitations}
Our experiments focus on mathematical reasoning benchmarks with relatively well-defined correctness criteria. It remains an open question whether the proposed framework generalizes to more open-ended generation tasks, such as dialogue or creative writing, where pseudo-label ambiguity may manifest differently.

To eliminate spurious signal amplification introduced by group-relative normalization, DDRL adopts a simple fixed, label-dependent advantage assignment. While this design is effective and stabilizes optimization in unsupervised settings, it represents a deliberately conservative choice. More expressive advantage formulations (e.g., confidence-adaptive scaling or uncertainty-aware advantage shaping) may further improve learning efficiency and robustness. We leave the exploration of more sophisticated advantage designs to future work.
\section*{Acknowledgements}
This research is supported by the National Natural Science Foundation of China under Grants-(62276256, U2441251). We thank Dong Yan at NLPR for early discussions and feedback on this project. Besides, we also extend our sincere thanks to the anonymous reviewers for their constructive suggestions.
We thank Meituan for providing academic exchange and hardware support in this work.
\bibliography{custom}

\input{sections/appendix}

\end{document}

%% file: sections/abs.tex
\begin{abstract}

Test-time reinforcement learning (TTRL) always adapts models at inference time via pseudo-labeling, leaving it vulnerable to spurious optimization signals from label noise.
Through an empirical study, we observe that responses with medium consistency form an ambiguity region and constitute the primary source of reward noise.
Crucially, we find that such spurious signals can be even amplified through group-relative advantage estimation.
Motivated by these findings, we propose a unified framework, Debiased and Denoised test-time Reinforcement Learning (DDRL), to mitigate spurious signals.
Concretely, DDRL first applies a frequency-based sampling strategy to exclude ambiguous samples while maintaining a balanced set of positive and negative examples.
It then adopts a debiased advantage estimation with fixed advantages, removing the bias introduced by group-relative policy optimization.
Finally, DDRL incorporates a consensus-based off-policy refinement stage, which leverages the rejection-sampled dataset to enable efficient and stable model updates.
Experiments on three large language models across multiple mathematical reasoning benchmarks demonstrate that DDRL consistently outperforms existing TTRL baselines.
The code will soon be released at \url{https://github.com/yuyongcan/DDRL}.
\end{abstract}

%% file: sections/intro.tex
\section{Introduction}

Reinforcement learning with verifiable rewards (RLVR) has recently emerged as an effective paradigm for improving large language models (LLMs) on structured challenging reasoning tasks, including mathematics and code generation~\citep{lambert2024tulu,yue2025does,jaech2024openai,guo2025deepseek,yu2025dapo,chen2025acereason,yu2025reassessing}.
By relying on explicit supervision or rule-based verification, RLVR enables stable optimization and strong task-specific performance.
However, its applicability is fundamentally constrained by the availability of reliable labels or verifiable reward functions, which limits its applicability when reliable labels or verifiers are unavailable, especially under distribution shift.

\begin{figure}[tbp]
    \centering
    \includegraphics[width=\linewidth]{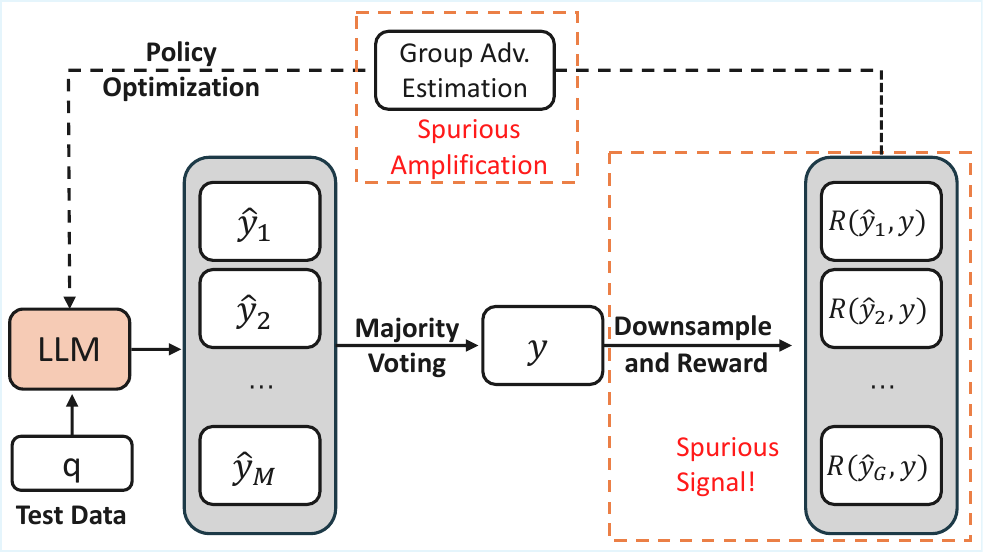}
    \caption{Overview of test-time reinforcement learning (TTRL). The spurious signal is generated during the reward stage with noisy pseudo-labels and amplified in the subsequent group-relative advantage estimation.}
    \label{fig:preliminary}
\end{figure}

To address distribution shifts during inference, Test-Time Reinforcement Learning (TTRL)~\citep{zuo2025ttrl} has emerged as a promising solution.
TTRL integrates test-time scaling (TTS)~\citep{muennighoff2025s1,zhang2025survey} with test-time training (TTT)~\citep{sun2020test}, enabling parameter updates via unsupervised reinforcement learning (RL).
As illustrated in \Cref{fig:preliminary}, TTRL generates multiple responses for each test query, derives a pseudo-label through majority voting, and optimizes the model using GRPO~\citep{shao2024deepseekmath} based on these pseudo-labels.
Despite its conceptual appeal, TTRL operates in a fundamentally unsupervised regime, where reward signals are derived entirely from the model’s own outputs.
This design renders TTRL highly vulnerable to spurious reward signals: incorrect responses may inadvertently receive positive rewards, while correct answers might be penalized, thereby distorting learning dynamics.

In this work, we systematically analyze the origins of spurious signals and how they propagate through the optimization process.
Empirically, we observe a strong correlation between answer frequency and reliability: high-frequency answers are predominantly correct, low-frequency answers are mostly incorrect, while answers with medium sampling frequency tend to be ambiguous and unreliable.
These medium-frequency responses constitute a major source of spurious reward signals.
Theoretically, we further demonstrate that GRPO’s advantage estimation introduces a systematic bias in this unsupervised setting. Specifically, its normalization mechanism disproportionately amplifies spurious rewards in low-consensus regimes.

Motivated by these insights, we propose \textbf{D}ebiased and \textbf{D}enoised test-time \textbf{R}einforcement \textbf{L}earning (DDRL), a framework designed to mitigate spurious signals.
First, we introduce a balanced confidence-aware sampling strategy that selects rollout samples based on their reliability while maintaining a balanced set of positive and negative examples.
Second, we replace group-relative advantage estimation with a bias-corrected scheme that assigns fixed, label-dependent advantages to eliminate the amplification effect.
Finally, we incorporate a consensus-based off-policy refinement stage after the RL phase, where a rejection sampling~\citep{zhang2023cumulative} dataset is constructed to enable efficient and stable post-RL optimization.

Demonstrated by experiments on multiple mathematical benchmarks and several LLMs, DDRL achieves significant relative improvements over TTRL, with gains of 15.3\% on Qwen2.5-MATH-1.5B and 12.7\% on LLaMA-3.1-8B-Instruct.
Our contributions are summarized as follows:
\begin{itemize}
    \item We reveal that medium-frequency samples are particularly prone to inducing noisy rewards. Furthermore, we find that group-relative advantage normalization used in TTRL will amplify these spurious signals subsequently.
    \item Based on the above findings, we propose DDRL, a framework that consists of a balanced confidence-aware sampling, debiased advantage estimation, and a lightweight consensus-based off-policy refinement to mitigate the spurious signals.
    \item Extensive experiments on several LLMs and several challenging mathematical benchmarks demonstrate the effectiveness of DDRL in mitigating spurious signals.
\end{itemize}

\begin{figure*}[htbp]
    \centering
    \begin{subfigure}[t]{0.4\linewidth}
        \centering
        \includegraphics[width=\linewidth]{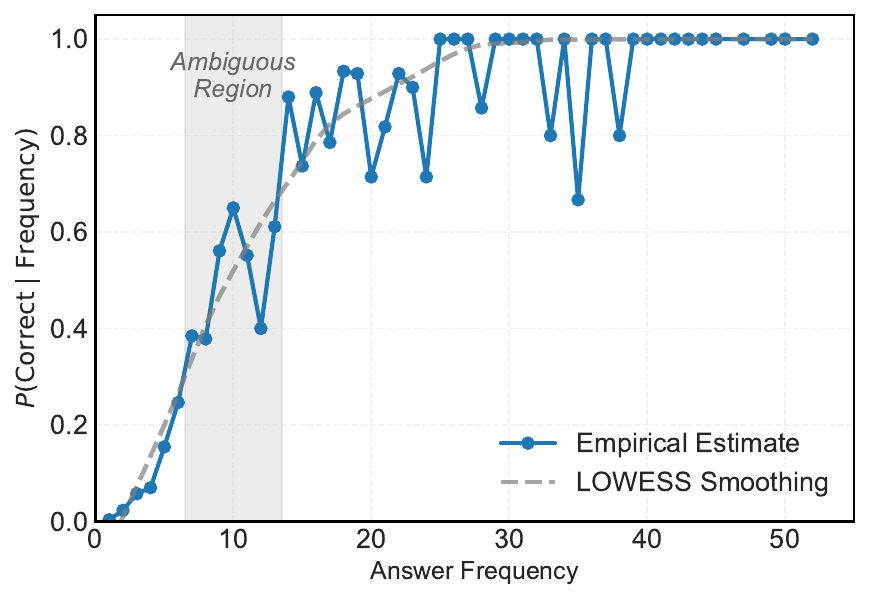}
        \caption{Correctness vs. frequency on MATH-500 (Qwen2.5-Math-1.5B).}
        \label{fig:freq_emp}
    \end{subfigure}
    \hspace{25pt}
    \begin{subfigure}[t]{0.4\linewidth}
        \centering
        \includegraphics[width=\linewidth]{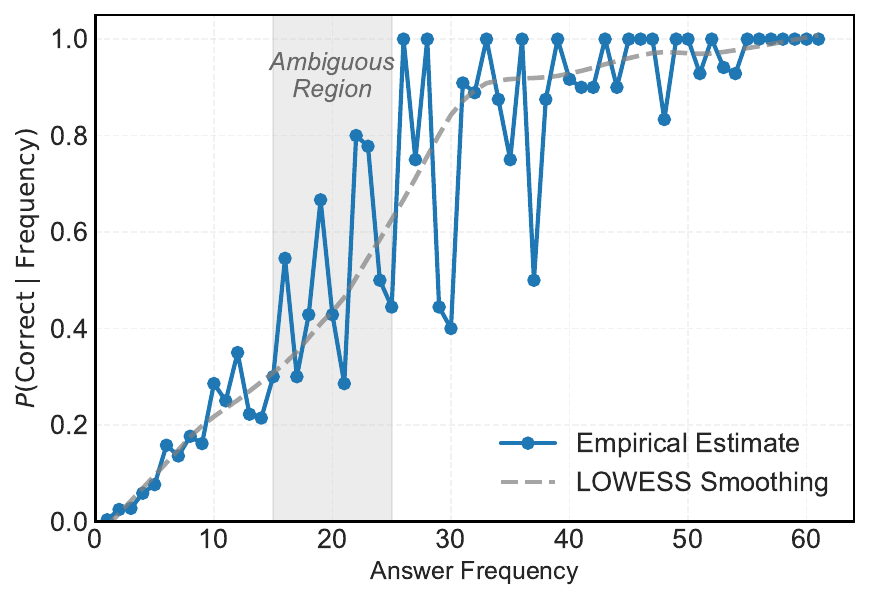}
        \caption{Correctness vs. frequency on MATH-500 (Qwen2.5-3B).}
        \label{fig:freq_the}
    \end{subfigure}
    \caption{Answer frequency as an imperfect proxy for reliability. We analyze the relationship between answer sampling frequency and correctness by sampling each prompt 64 times and grouping answers by frequency. High-frequency answers are generally correct, while low-frequency answers are mostly incorrect. In contrast, answers with medium sampling frequency exhibit high variance in correctness (shaded region), forming an ambiguous regime where it is hard to determine the answer correctness. These ambiguous samples constitute a major source of spurious reward signals in TTRL.
}
    \label{fig:freq}
\end{figure*}

%% file: sections/observations.tex
\section{Understanding Spurious Signals}
\label{sec:obs}
\subsection{Test-Time Reinforcement Learning}
Test-time reinforcement learning (TTRL)~\citep{zuo2025ttrl} adapts a pre-trained language model $\pi_\theta$ at inference time via RL, specifically avoiding reliance on ground-truth supervision.
Instead, TTRL constructs training signals intrinsically using test-time scaling.

Given an unlabeled test query $q$, the model samples $N$ candidate responses $\{y_1, y_2, \dots, y_N\}$ by repeated sampling $y_i \sim \pi_\theta(\cdot \mid q)$.
A consensus pseudo-label $y^*$ is then derived via majority voting over the sampled responses.
Based on this pseudo-label, TTRL defines a binary reward function
\begin{equation}
    r(y, y^*) = \mathbb{I}(y = y^*),
\end{equation}
and seeks to maximize the expected reward under the current policy:
\begin{equation}
    \max_\theta \; \mathbb{E}_{y \sim \pi_\theta(\cdot \mid q)} \big[ r(y, y^*) \big].
\end{equation}
Notably, both the reward signal and the optimization target are induced entirely from model-generated outputs, without external verification.

To optimize $\pi_\theta$ using the sampled rollouts, TTRL adopts GRPO ~\citep{shao2024deepseekmath} as the underlying reinforcement learning algorithm.
GRPO computes advantages by normalizing rewards within each sampled group:
\begin{equation}
    A_i = \frac{r_i - \mathrm{mean}(r)}{\mathrm{std}(r)}.
\end{equation}
The policy is then updated by maximizing the clipped surrogate objective:
\begin{equation}
\begin{aligned}
    \max_\theta \; \mathbb{E}_{i}\Big[
    \min\big( & f_i(\theta) A_i, \\
    & \mathrm{clip}(f_i(\theta), 1-\epsilon, 1+\epsilon) A_i\big)
    \Big]
\end{aligned}
\end{equation}
where $f_i(\theta) = \frac{\pi_\theta(y_i \mid q)}{\pi_{\theta_{\mathrm{old}}}(y_i \mid q)}$ and $\epsilon$ denotes the clipping threshold.

\subsection{Origin of Spurious Reward Signals through Answer Frequency}
\label{sec:obs_1}
Majority voting provides a convenient mechanism for constructing pseudo-labels at test time, but it does not guarantee alignment with ground-truth labels.
As a result, spurious reward signals are often unavoidable when all sampled rollouts are treated equally during optimization.
To better understand when pseudo-labels are reliable, we analyze the relationship between answer correctness and sampling frequency.

As shown in \Cref{fig:freq}, the correctness of a sampled response is strongly correlated with how frequently it appears among repeated generations.
High-frequency responses are predominantly correct, while low-frequency responses are mostly incorrect.
In contrast, responses with medium sampling frequency exhibit substantial uncertainty: their correctness probability varies sharply and is highly unstable.
We provide a Bayesian explanation for this phenomenon in the Appendix~\ref{sec:prob}.

These medium-frequency samples constitute a dominant source of spurious reward signals, as they are neither consistently correct nor consistently incorrect.
Therefore, ideally, we should minimize the use of these samples to reduce spurious signals.
However, standard TTRL treats all sampled rollouts the same in optimization, allowing ambiguous medium-frequency samples to exert a disproportionate influence on the learning signal.

\subsection{Amplification of Spurious Signals by Group-Relative Advantage Estimation}
\label{sec:obs_2}

\begin{figure*}[htbp]
    \centering
    \begin{subfigure}[t]{0.4\linewidth}
        \centering
        \includegraphics[width=\linewidth]{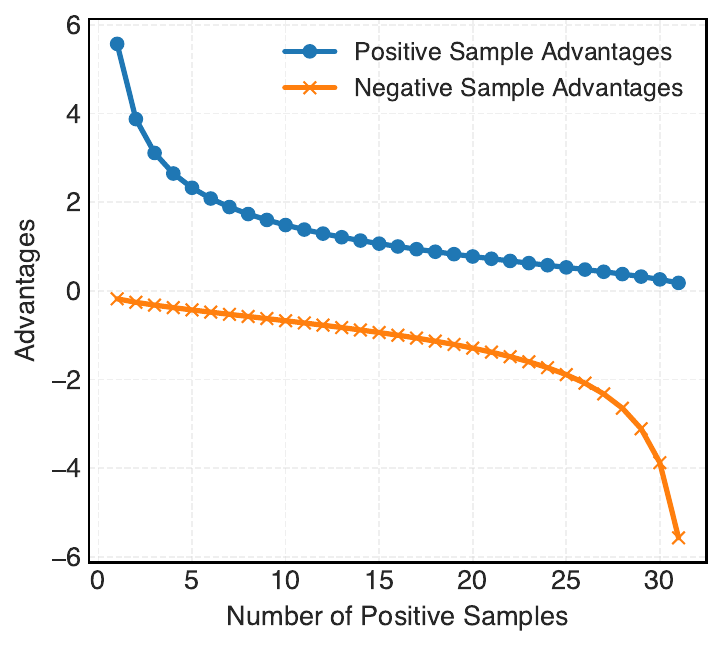}
        \caption{Advantage vs. number of positive samples in GRPO under 32 rollouts. When positive samples are
scarce, normalization over group statistics
yields large advantage magnitudes.
}
        \label{fig:GRPO_adv}
    \end{subfigure}
    \hfill
    \begin{subfigure}[t]{0.55\linewidth}
        \centering
        \includegraphics[width=\linewidth]{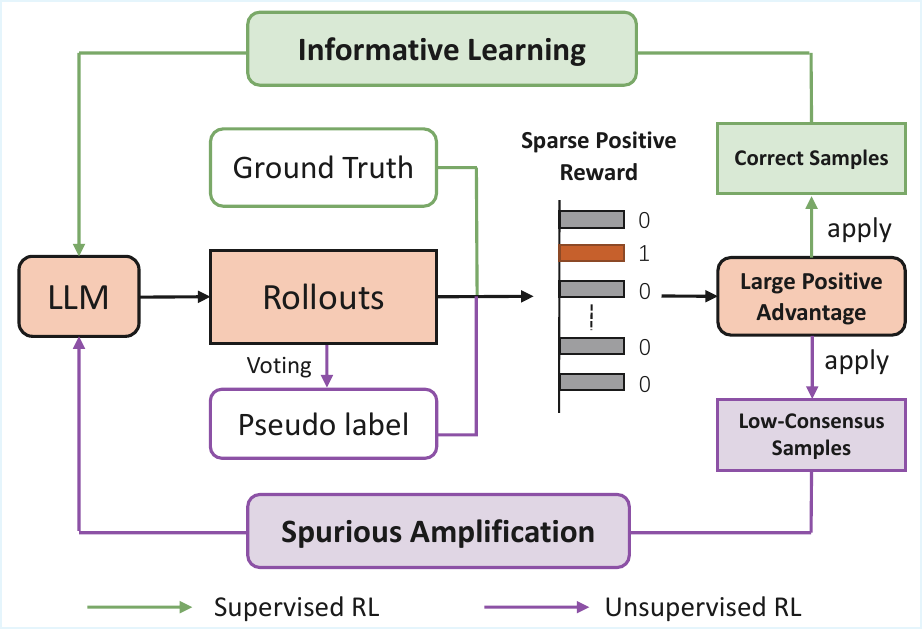}
        \caption{Conceptual comparison between supervised and unsupervised settings. In supervised RL, sparse positive rewards correspond to rare but reliable signals. In unsupervised TTRL, sparse positive samples indicate low consensus and unreliable pseudo-labels; assigning large advantages to ambiguous samples amplifies spurious signals.}
        \label{fig:GRPO_URL}
    \end{subfigure}
    \caption{Behavior of group-relative advantage estimation under limited positive samples.
}
    \label{fig:GRPO_bias}
\end{figure*}

GRPO~\citep{shao2024deepseekmath} has proven highly effective in supervised RL, particularly in settings where reliable labels ensure that positive rewards correspond to correct behaviors.
However, its inductive bias implicitly assumes that rare positive samples are informative and trustworthy.
In unsupervised RL, this assumption no longer holds.

As illustrated in \Cref{fig:GRPO_bias}, GRPO constructs advantages by normalizing rewards within each group of sampled rollouts.
When the number of positive samples is small, this normalization leads to large advantage magnitudes for positive samples.
In supervised scenarios, such behavior is desirable: ground-truth supervision guarantees that rare positive samples are correct, allowing the optimizer to emphasize informative signals.

In the context of TTRL, however, a small number of positive samples does not indicate scarcity of valuable supervision.
Instead, since rewards are derived from majority-voted pseudo-labels, few positive samples directly reflect low consensus among model-generated responses.
Low consensus, in turn, implies high uncertainty about the correctness of the pseudo-label itself.
As a result, GRPO assigns large advantages precisely to those samples whose labels are the least reliable.

\begin{table}[htbp]
\centering
\resizebox{.8\linewidth}{!}{
\begin{tabular}{l|cc} \toprule
Adv. Estimation & AIME2024 & MATH \\ \midrule
GRPO & 15.8 & 73.0  \\
GRPO (w/o norm.) &  20.6& 75.0 \\ \bottomrule
\end{tabular}
}
\caption{A preliminary comparison between group-relative and fixed advantage estimation in TTRL using Qwen2.5-Math-1.5B. GRPO (w/o norm.) removes the advantage normalization in GRPO.}
\label{tab:adv_cmp}
\end{table}

This mismatch between GRPO’s inductive bias and the semantics of pseudo-labels causes low-consensus and potentially incorrect samples to exert a disproportionate influence on policy updates.
Consequently, spurious reward signals introduced during pseudo-label construction are further amplified during advantage estimation, leading to unstable and misleading optimization dynamics.

To isolate the effect of advantage normalization, we conduct a preliminary comparison between GRPO and a simple fixed advantage scheme.
As shown in \Cref{tab:adv_cmp}, replacing group-relative advantages with constant, label-dependent values already yields consistent improvements across benchmarks.
This observation suggests that the instability arises primarily from inappropriate advantage scaling.

%% file: sections/method.tex
\section{Mitigating Spurious Signals}
Based on the analysis in \Cref{sec:obs}, we propose Debiased and Denoised Test-Time Reinforcement Learning (DDRL) to mitigate the spurious signals.

\subsection{Balanced Confidence-Aware Sampling}

To denoise the potential spurious signal brought by inappropriate reward, we propose \emph{balanced confidence-aware sampling}, which selectively filters noisy samples while preserving informative supervision.
Our design is guided by two principles: (i) confidence-aware selection based on sampling frequency, and (ii) balanced exposure to positive and negative samples.

Formally, let $c(y_i)$ denote the occurrence count of response $y_i$ among $N$ sampled rollouts.
We use $c(y_i)$ as a confidence indicator, motivated by the empirical observation in \Cref{sec:obs_1} that the conditional correctness probability $P(\mathrm{correct} \mid c)$ is high at extreme values of $c$ and highly unstable in the medium-frequency region.
Rather than optimizing over all rollouts, we select a fixed number $K$ of samples for each prompt.

We first determine the number of positive samples as:
\begin{equation}
    K^+ = \min\!\left(c(y^*),\; \left\lfloor \frac{K}{2} \right\rfloor \right),
\end{equation}
where $y^*$ is the pseudo-label and $\lfloor \cdot \rfloor$ denotes the floor function.
This formulation ensures that (1) we do not select more positives than are available, and (2) positive samples never dominate the batch (capped at $50\%$), thereby enforcing a balanced label distribution.
The number of negative samples is then given by:
\begin{equation}
    K^- = K - K^+.
\end{equation}
Based on these quotas, the sampling procedure is defined as follows.
\textbf{Positive selection}: we select the top-$K^+$ samples corresponding to the pseudo-label $y^*$.
\textbf{Negative selection}: we select the $K^-$ samples with the \textit{lowest} occurrence counts.
We treat low-frequency samples as negatives to \textit{mitigate false negatives}. In complex reasoning tasks, high-frequency alternatives may correspond to valid or even correct reasoning paths and should not be penalized. In contrast, rare outliers are statistically more likely to be incorrect hallucinations, making them safe negatives. By discarding medium-frequency responses, we remove the high-variance ambiguous region from optimization.

\subsection{Debiased Advantage Estimation}
Motivated by the analysis in \Cref{sec:obs_2}, to mitigate the spurious amplification introduced by GRPO advantage estimation, we replace group-relative advantage estimation with a bias-corrected, fixed advantage assignment for rollout $y_i$:
\begin{equation}
    A_i = \mathbb{I}(y=y^*)-\mathbb{I}(y\neq y^*),
\end{equation}
where positive samples receive a constant advantage of $+1$ and negative samples receive $-1$, independent of the number of positive samples in the group.
By decoupling advantage magnitude from group statistics, this formulation eliminates the amplification effect induced by normalization, resulting in more stable and reliable optimization under unsupervised pseudo-labels.

\subsection{Consensus-Based Off-Policy Refinement}

Although the preceding components reduce spurious signals during on-policy RL, policy updates can still be noisy due to stochastic optimization.
However, the consensus here is typically much stronger than in the initial phase.
We therefore introduce a lightweight off-policy refinement stage to consolidate the improvements.

Let $\pi_{\theta_{\mathrm{RL}}}$ denote the policy model after the RL phase.
For each test query $q \in \mathcal{Q}$, we sample $M$ responses as follows:
\begin{equation}
    \{y_1, \dots, y_M\}, \quad y_j \sim \pi_{\theta_{\mathrm{RL}}}(\cdot \mid q),
\end{equation}
with $M=128$ in our experiments.
A consensus pseudo-label $y^*(q)$ is obtained via majority voting, and rejection sampling is applied to retain only responses that agree with the consensus:
\begin{equation}
    \mathcal{A}(q) = \{(q, y_j) \mid y_j = y^*(q)\}.
\end{equation}
We then aggregate the accepted samples and yield the off-policy dataset:
\begin{equation}
    \mathcal{D}_{\mathrm{op}} = \bigcup_{q \in \mathcal{Q}} \mathcal{A}(q).
\end{equation}

Starting from $\theta_{\mathrm{RL}}$, we perform supervised fine-tuning by maximizing
\begin{equation}
    \mathbb{E}_{(q,y)\sim\mathcal{D}_{\mathrm{op}}}[\log \pi_\theta(y\mid q)].
\end{equation}
This stage distills high-consensus behaviors of the RL-adapted policy into the model, providing a stable and efficient refinement of test-time updates.

%% file: sections/exp.tex
\section{Experiments}
\subsection{Setup}

\input{tabs/main_res}
\textbf{Benchmarks and Models.}
We evaluate DDRL on three widely used mathematical reasoning benchmarks: AIME 2024~\citep{li2024numinamath}, AMC~\citep{li2024numinamath}, and MATH-500~\citep{hendrycks2measuring}.
To assess robustness across model architectures and scales, we conduct experiments on a diverse set of models, including a base LLM (Qwen2.5-3B~\citep{qwen2025qwen25technicalreport}), an instruction-tuned LLM (Llama-3.1-8B-Instruct~\citep{grattafiori2024llama}), and a math-specialized model (Qwen2.5-Math-1.5B~\citep{yang2024qwen2}).
\input{tabs/abla}
\textbf{Baselines.}
We compare DDRL against representative test-time training methods.
Specifically, we include \textbf{TTRL}~\citep{zuo2025ttrl}, which performs unsupervised reinforcement learning using majority-vote pseudo-labels, and \textbf{ETMR}~\citep{liu2025ettrl}, which improves rollout diversity by selectively forking trajectories at high-entropy tokens.
We also report results of the original pretrained models without any test-time adaptation, referred to as \textbf{No Adaptation}.

\textbf{Implementation Details and Evaluation Protocol.}
All experiments are implemented using the \texttt{verl} framework~\citep{sheng2024hybridflow}.
For pseudo-label estimation, we sample $N=64$ rollouts per prompt with temperature $1.0$ for math-specialized models and $0.6$ for other models, following prior work~\citep{zuo2025ttrl}.
These rollouts are then downsampled to $32$ trajectories for training.
We optimize models using AdamW with a cosine learning rate schedule, where the peak learning rate is set to $5\times10^{-7}$ for Qwen models and $2\times10^{-7}$ for the Llama model to account for scale differences.
The maximum generation length is fixed to $3072$ tokens.
The number of training episodes is set to $10$, $30$, and $80$ for MATH-500, AMC, and AIME 2024, respectively, proportional to dataset size.
In the off-policy refinement stage, we apply rejection sampling independently for each prompt and retain a fixed number of high-confidence samples (4 for MATH-500 and 16 for AMC and AIME 2024).
We then perform supervised fine-tuning for 5 epochs with a learning rate of $1\times10^{-5}$.
For evaluation, we report pass@1 following DeepSeek-R1, where $16$ responses are generated per prompt with temperature $0.6$ and top-$p=0.95$, and accuracy is averaged across prompts.

\subsection{Main Results}

\Cref{tab:main} compares DDRL with baseline methods using pass@1 across three models and multiple mathematical reasoning benchmarks.

Across all settings, DDRL consistently improves over TTRL and, in most cases, over the stronger ETMR baseline.
On Qwen2.5-Math-1.5B, DDRL achieves substantial gains over ETMR, improving performance by $+19.0\%$ on AIME 2024, $+4.1\%$ on AMC, and $+4.9\%$ on MATH-500, which corresponds to a $+6.7\%$ average improvement.
These results indicate that mitigating spurious reward signals is particularly effective when the base model already exhibits strong reasoning ability.

On Qwen2.5-Base-3B, DDRL yields consistent but more moderate improvements across all benchmarks, resulting in a $+1.7\%$ average gain over ETMR.
This suggests that while DDRL is broadly effective, the magnitude of improvement depends on the capacity of the underlying model.

On LLaMA-3.1-8B-Instruct, DDRL substantially improves performance on AMC and MATH-500, with gains of $+9.0\%$ and $+13.6\%$, respectively.
Although DDRL underperforms ETMR on AIME 2024, it remains consistently stronger than TTRL and achieves the best average performance among all methods.

\begin{figure*}[htbp]
    \centering
    \begin{subfigure}[t]{0.46\linewidth}
        \centering
        \includegraphics[width=\linewidth]{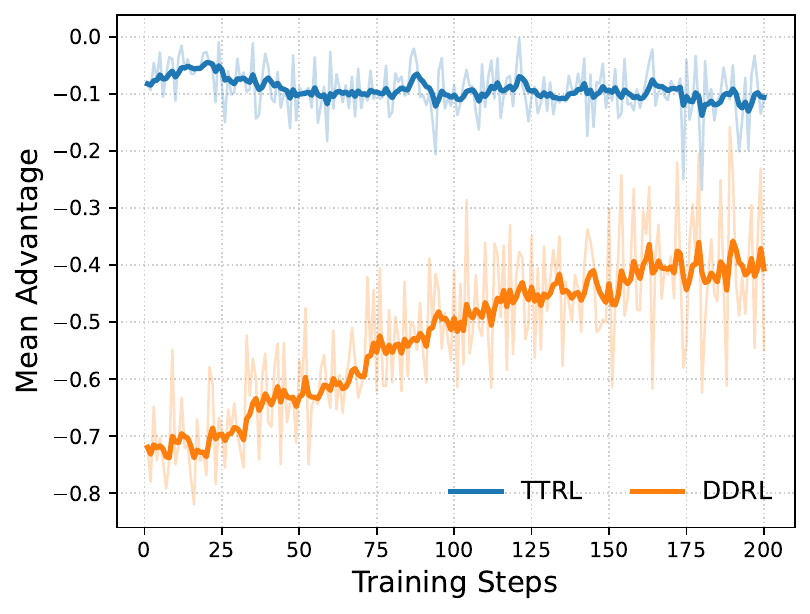}
        \caption{AIME 2024 with Qwen2.5-Math-1.5B.}
        \label{fig:adv_dyn_aime}
    \end{subfigure}
    \hfill
    \begin{subfigure}[t]{0.46\linewidth}
        \centering
        \includegraphics[width=\linewidth]{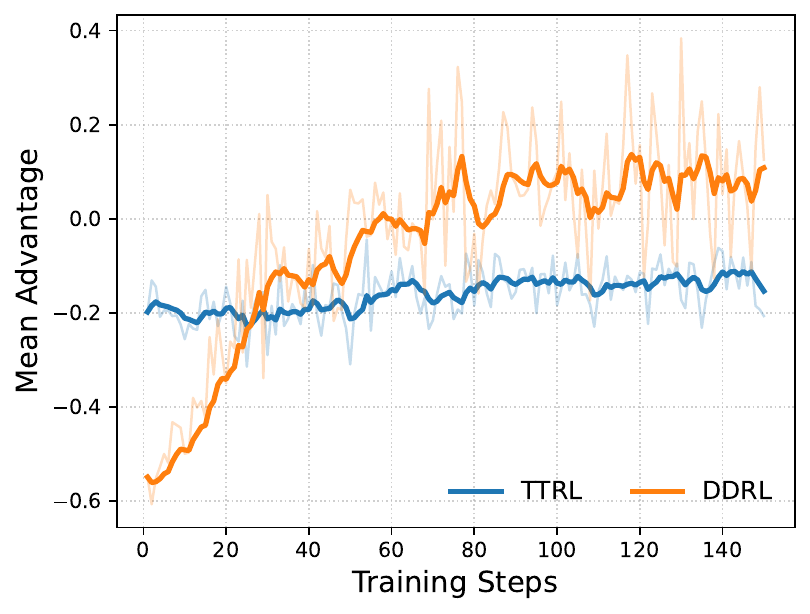}
        \caption{MATH-500 with Qwen2.5-Math-1.5B.}
        \label{fig:adv_dyn_math}
    \end{subfigure}
    \caption{Comparison of training dynamics between TTRL and DDRL via mean advantage.}
    \label{fig:adv_dyn}
\end{figure*}

\subsection{Ablation Study}

\Cref{tab:abla} reports an ablation study analyzing the individual and combined effects of the proposed components in DDRL.

Applying balanced confidence-aware sampling (BCS) alone improves performance on Qwen2.5-Math-1.5B and LLaMA-3.1-8B-Instruct, indicating that filtering medium-frequency, ambiguous samples effectively suppresses spurious reward signals.
However, on Qwen2.5-Base-3B, BCS alone leads to a slight degradation, suggesting that reducing noisy supervision without correcting the optimization dynamics can limit learning for some models.

Incorporating debiased advantage estimation (DAE) consistently improves performance across all models.
Compared to BCS-only training, DAE stabilizes optimization by removing frequency-dependent advantage amplification, recovering the performance drop on Qwen2.5-Base-3B and yielding clear gains on the other models.
This demonstrates that debiasing advantage estimation is essential for robust learning under unsupervised reinforcement learning with pseudo-labels.

Adding consensus-based off-policy refinement (COR) further improves results in all settings.
The largest gains are observed on Qwen2.5-Math-1.5B, where high-consensus behaviors learned during reinforcement learning can be effectively consolidated.
On Qwen2.5-Base-3B and LLaMA-3.1-8B-Instruct, COR provides smaller but consistent improvements, indicating that this stage primarily serves as a stabilization and refinement step rather than a primary driver of performance.

\subsection{Additional Analysis}

\textbf{Advantage Dynamics.}
\Cref{fig:adv_dyn} compares the training dynamics of TTRL and DDRL by tracking the mean advantage during optimization on AIME 2024 and MATH-500.
Since the advantage reflects the relative contribution of positive and negative samples, its magnitude and sign provide insight into how learning signals evolve over training.

On AIME 2024, the mean advantage under TTRL remains close to zero throughout training, indicating limited differentiation between positive and negative samples' contributions.
In contrast, DDRL maintains a consistently negative mean advantage, suggesting that optimization is dominated by negative samples.
Given the high difficulty of AIME problems and the low reliability of early pseudo-labels, this behavior suppresses premature reinforcement of incorrect positive signals.

On MATH-500, DDRL exhibits a distinct transition in advantage dynamics.
Early in training, the mean advantage is strongly negative, reflecting conservative learning when pseudo-label confidence is low.
As training progresses, the mean advantage gradually increases and becomes positive, indicating a shift toward learning from positive samples as pseudo-label quality improves.
This adaptive transition is not observed in TTRL.



\input{tabs/sft_cost}
\textbf{Analysis of the Off-Policy Stage.}
To assess the efficiency and effectiveness of the consensus-based off-policy refinement stage, we compare it with extending the on-policy RL phase under comparable computational budgets.
Experiments are conducted on Qwen2.5-MATH-1.5B by allocating additional RL training time corresponding to approximately one-tenth of the original RL budget for each dataset.

As reported in \Cref{tab:offpolicy_vs_rl}, additional on-policy RL yields limited or inconsistent improvements and can even degrade performance suggested on AIME 2024.
In contrast, the off-policy refinement stage consistently achieves larger and more stable gains with substantially lower training cost.
Training for only five SFT epochs (3-5 minutes) improves performance across all datasets, outperforming extended RL runs that require up to 72 minutes.
These results indicate that consolidating high-consensus behaviors via off-policy supervision is a more efficient and reliable alternative to prolonging on-policy reinforcement learning.

%% file: tabs/main_res.tex
\begin{table*}[htbp]
\begin{center}
\resizebox{0.8\textwidth}{!}{
\begin{tabular}{llcccc}
\toprule
Model                              & Name      & AIME 2024  & AMC   & MATH-500  & Avg   \\ \midrule
\multirow{4}{*}{Qwen2.5-Math-1.5B} 
& No Adaptation & 7.7 & 28.6 & 32.7 & 23.0 \\
& TTRL~\citep{zuo2025ttrl}     & 15.8       & 48.9  & 73.0      & 45.9  \\
                                   & ETMR~\citep{liu2025ettrl}      & 21.0       & 50.8  & 76.9      & 49.6  \\
                       & \cellcolor{headerblue}DDRL      & \cellcolor{headerblue}\textbf{25.0}       &\cellcolor{headerblue}\textbf{52.9}  &\cellcolor{headerblue}\textbf{80.7}      &\cellcolor{headerblue}\textbf{52.9}  \\  		 
                                   &\cellcolor{headerblue}$\Delta$  &\cellcolor{headerblue}\color{red}{$\uparrow$19.0\%}  &\cellcolor{headerblue}\color{red}{$\uparrow$4.1\%}  &\cellcolor{headerblue}\color{red}{$\uparrow$4.9\%}  &\cellcolor{headerblue}\color{red}{$\uparrow$6.7\%} \\ \midrule
\multirow{4}{*}{Qwen2.5-Base-3B}
& No Adaptation &4.4  &24.5 &53.2 &27.4 \\
& TTRL~\citep{zuo2025ttrl}     & 7.9        & 40.7  & 72.2      & 40.3  \\
                                   & ETMR~\citep{liu2025ettrl}     & 9.2        & 41.7  & 71.7      & 40.9  \\ 			 
                                   &\cellcolor{headerblue}DDRL      &\cellcolor{headerblue}\textbf{10.0}       &\cellcolor{headerblue}\textbf{42.2}  &\cellcolor{headerblue}\textbf{72.7}      &\cellcolor{headerblue}\textbf{41.6}  \\ 
                                   &\cellcolor{headerblue}$\Delta$  &\cellcolor{headerblue}\color{red}{$\uparrow$8.2\%}  &\cellcolor{headerblue}\color{red}{$\uparrow$1.2\%}  &\cellcolor{headerblue}\color{red}{$\uparrow$1.4\%}  &\cellcolor{headerblue}\color{red}{$\uparrow$1.7\%} \\ \midrule
\multirow{4}{*}{Llama-3.1-8B-Instruct}   
& No Adaptation & 4.6 & 23.3 & 48.6 & 25.5 \\
& TTRL~\citep{zuo2025ttrl}     & 10.0       & 32.3  & 63.7      & 35.3  \\
                                   & ETMR~\citep{liu2025ettrl}     & \textbf{16.9}       & 35.4  & 59.5      & 37.3  \\
                                   &\cellcolor{headerblue}DDRL      &\cellcolor{headerblue}13.3       &\cellcolor{headerblue}\textbf{38.6}  &\cellcolor{headerblue}\textbf{67.6}      &\cellcolor{headerblue}\textbf{39.8}  \\	
                                   &\cellcolor{headerblue}$\Delta$  &\cellcolor{headerblue}\color{green}{$\downarrow$21.3\%}  &\cellcolor{headerblue}\color{red}{$\uparrow$9.0\%}  &\cellcolor{headerblue}\color{red}{$\uparrow$13.6\%}  &\cellcolor{headerblue}\color{red}{$\uparrow$6.7\%}    \\
\bottomrule
\end{tabular}
}
\caption{Performance (pass@1) comparison among DDRL and baseline methods. $\Delta$ represents the percentage performance gain achieved by our method compared to ETMR. The best results are in bold.}
\label{tab:main}
\end{center}
\end{table*}

%% file: tabs/abla.tex
\begin{table*}[ht]
\begin{center}
\resizebox{0.8\textwidth}{!}{
\begin{tabular}{lccccccc}
\toprule
Model                              & BCS   & DAE     & COR       & AIME 2024  & AMC   & MATH-500  & Avg   \\ \midrule
\multirow{4}{*}{Qwen2.5-Math-1.5B} & -          & -                  & -         & 15.8       & 48.9  & 73.0      & 45.9   \\	
                                   & \cmark     & -                  & -         & 17.3       & 46.9  & 74.1      & 46.1  \\				 
                                   & \cmark     & \cmark             & -         & 20.2       & 46.3  & 74.6      & 47.0  \\
                                   & \cmark     & \cmark             & \cmark    & 25.0       & 52.9  & 80.7      & 52.9  \\  \midrule
\multirow{3}{*}{Qwen2.5-Base-3B}   & -          & -                  & -         & 7.9        & 40.7  & 72.2      & 40.3  \\ 				 
                                   & \cmark     & -                  & -         & 6.7        & 41.3  & 72.0      & 40.0  \\
                                   & \cmark     & \cmark             & -         & 10.0       & 42.2  & 72.3      & 41.5  \\				 
                                   & \cmark     & \cmark             & \cmark    & 10.0       & 42.2  & 72.7      & 41.6  \\  \midrule
\multirow{3}{*}{Llama-3.1-8B-Instruct}      & -          & -                  & -         & 10.0       & 32.3  & 63.7      & 35.3  \\	 			 
                                   & \cmark     & -                  & -         & 13.3       & 38.4  & 64.4      & 38.7  \\
                                   & \cmark     & \cmark             & -         & 13.3       & 38.5  & 67.4      & 39.7  \\
                                   & \cmark     & \cmark             & \cmark    & 13.3       & 38.6  & 67.6      & 39.8  \\	
\bottomrule
\end{tabular}
}
\caption{Ablation study of DDRL.
BCS, DAE, and COR denote balanced confidence-aware sampling, debiased advantage estimation, and consensus-based off-policy refinement, respectively.
The results demonstrate the complementary effects of reducing spurious signals, correcting advantage bias, and consolidating learned behaviors.}
\label{tab:abla}
\end{center}
\end{table*}

%% file: tabs/sft_cost.tex
\begin{table}[h]
\centering
\resizebox{\linewidth}{!}{
\begin{tabular}{l l c c c}
\toprule
Dataset & Setting & Epoch & Time ($\downarrow$) & Pass@1 ($\uparrow$) \\
\midrule
\multirow{3}{*}{AIME}
 & BCS + DAE & -- & -- & 20.2 \\
 & + Additional RL & 8 & 15 min & 19.2 \\
 & + COR (Ours) & 5 & \textbf{3 min} & \textbf{25.0} \\
\midrule
\multirow{3}{*}{AMC}
 & BCS + DAE & -- & -- & 46.3 \\
 & + Additional RL & 3 & 42 min & 46.9 \\
 & + COR (Ours) & 5 & \textbf{3 min} & \textbf{52.9} \\
\midrule
\multirow{3}{*}{MATH}
 & BCS + DAE & -- & -- & 74.6 \\
 & + Additional RL & 1 & 72 min & 74.4 \\
 & + COR (Ours) & 5 & \textbf{5 min} & \textbf{80.7} \\
\bottomrule
\end{tabular}
}
\caption{Comparison between additional on-policy RL training and the consensus-based off-policy refinement stage on Qwen2.5-MATH-1.5B. ``BCS + DAE'' denotes the model performance after the RL stage without any additional training.}
\label{tab:offpolicy_vs_rl}
\end{table}

%% file: sections/rel_work.tex
\section{Related Work}
\subsection{Test-Time Training}
Test-time training \citep{sun2020test,liang2025comprehensive,yu2025test} refers to a paradigm where a model is adapted during inference by updating its parameters using self-supervised or pseudo-supervised signals available at test time, without access to ground-truth labels.
In early test-time training (TTT) paradigms, pseudo-labeling ~\citep{liang2020we} and entropy minimization~\citep{wang2021tent,yu2024stamp,yu2023benchmarking} are the primary techniques used to enhance a model’s generalization ability under distribution shift during inference-time.

In the context of LLMs, LMSI~\citep{huang2023large} uses the high-confidence chain-of-thought (CoT) trajectories with majority answers to SFT the language model for improving both in-domain and out-of-domain performance.
Similarly, SEALONG~\citep{li2024large} scores the sampled trajectories with Minimum Bayes Risks and then applies SFT and preference optimization.
TLM~\citep{hu2025test} constructs a test-time learning paradigm for LLMs that adapts to domain shifts by minimizing the input perplexity of unlabeled test data, with sample-efficient selection and LoRA-based updates to mitigate instability.
TTRL~\citep{zuo2025ttrl} combines the TTT and TTS, using the majority voting answer to conduct RL on unlabeled data.

\subsection{Unsupervised Reinforcement Learning}
TTRL~\citep{zuo2025ttrl} proposes to perform reinforcement learning using unlabeled data.
Recently, a growing body of work has extended this paradigm ~\citep{prabhudesai2025maximizing,zhao2025learning,wu2025spine,zhang2025consistent,liu2025ettrl,yu2025restrain,wei2025unsupervised,yan2026if}.
MM-UPT~\citep{wei2025unsupervised} extends the TTRL to a multi-modal setting.
RENT~\citep{prabhudesai2025maximizing} and INTUITOR~\citep{zhao2025learning} both use the model's own confidence as the reward signal for RL.
ETTRL~\citep{liu2025ettrl} proposes entropy-fork tree majority rollout to fork rollout only at the token with high entropy to form a diverse set and reshape the advantage according to the entropy.
Similarly, SPINE~\citep{wu2025spine} forks the rollout only at high entropy tokens and only applies loss on the fork token.
RESTRAIN~\citep{yu2025restrain} applies a negative rollout penalization when the model's prediction is in a low consensus and weights each prompt with an offline estimated confidence.
In contrast to these approaches, which primarily focus on designing alternative rewards or rollout strategies, our work analyzes how spurious signals arise and are amplified during optimization in unsupervised settings.
We show that advantage estimation itself can introduce systematic bias when pseudo-label reliability is low, and propose a unified framework that jointly addresses the problem.

%% file: sections/conclu.tex
\section{Conclusion}
In this work, we investigate the prevailing spurious signals in TTRL. We first empirically demonstrate that ambiguous medium-frequency samples are the major source of spurious signals and group-relative advantage estimation amplifies these signals, leading to a misguided optimization. Based on these insights, we propose debiased and denoised test-time reinforcement learning, which denoises the spurious signals through a confidence-aware sampling and applies a debiased advantage estimation.
Moreover, a lightweight consensus-based off-policy refinement stage is introduced to further enhance the consensus. Extensive experiments on multiple language models and mathematical reasoning benchmarks demonstrate that DDRL consistently outperforms existing TTRL-based methods under comparable computational budgets. These results highlight the importance of controlling uncertainty and bias in test-time optimization and point to more robust directions for unsupervised reinforcement learning at inference time.


%% file: sections/appendix.tex
\appendix

\section{Implementation Detail on Input Prompt Format}
Following TTRL~\citep{zuo2025ttrl}, we add the reasoning prompt to the model:
\begin{systemprompt}
Please reason step by step, and put your
final answer within \boxed{}.
\end{systemprompt}
For Math-specialized models, we use it as a system prompt; for other models, we append the cue to the question.

\section{Why is DDRL Effective in Mitigating Spurious Signals?}
It is widely recognized that spurious learning signals in TTRL primarily originate from erroneous pseudo-labels. Since DDRL also relies on pseudo-labels for optimization, a natural question arises: how does DDRL mitigate spurious signals without eliminating pseudo-labeling altogether?

We argue that spurious signals introduced during the reward construction stage stem from two distinct sources: \emph{false negatives} and \emph{false positives}. DDRL is designed to explicitly suppress the impact of both.

\textbf{False Negatives.}  
False negatives correspond to correct answers that are mistakenly assigned negative rewards. DDRL mitigates this issue through balanced confidence-aware sampling, which prioritizes samples with extreme answer frequencies. As shown in \Cref{fig:freq}, low-frequency answers are overwhelmingly incorrect, while correct answers are more likely to appear at higher frequencies. By preferentially selecting low-frequency samples as negative examples, DDRL substantially reduces the probability of assigning negative rewards to potentially correct answers, thereby suppressing false negatives.

\begin{figure*}[htbp]
    \centering
    \begin{subfigure}[t]{0.4\linewidth}
        \centering
        \includegraphics[width=\linewidth]{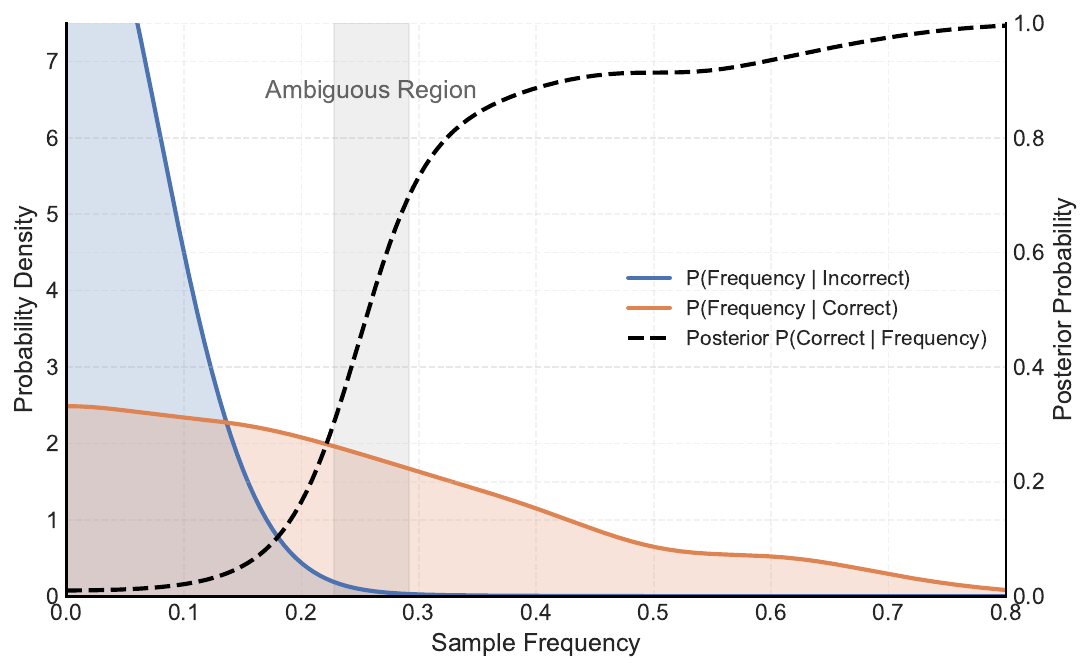}
        \caption{Answer distribution on MATH-500 with Qwen2.5-Math-1.5B.}
        \label{fig:dist_math}
    \end{subfigure}
    \hspace{25pt}
    \begin{subfigure}[t]{0.4\linewidth}
        \centering
        \includegraphics[width=\linewidth]{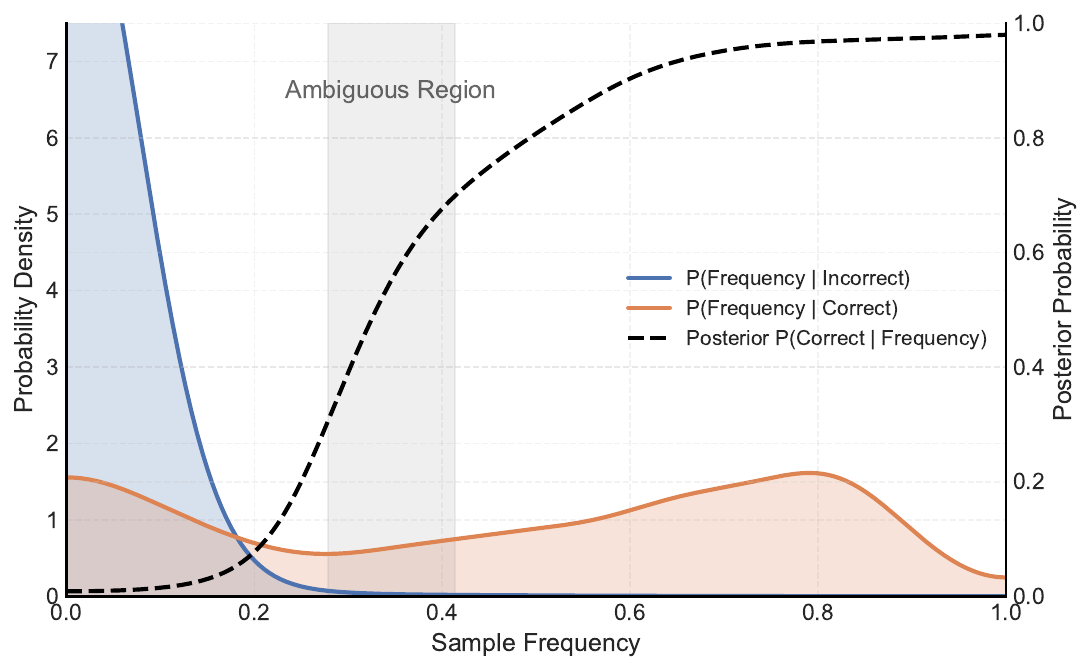}
        \caption{Answer distribution on MATH-500 with Qwen2.5-3B.}
        \label{fig:dist_qwen}
    \end{subfigure}
    \caption{Conditional answer distributions reveal ambiguous frequency regions in self-consistency in the \Cref{fig:freq}.
}
 \label{fig:dist}
\end{figure*}

\textbf{False Positives.}  
False positives arise when incorrect answers are incorrectly assigned positive rewards due to unreliable pseudo-labels. Although such samples cannot be entirely avoided, DDRL limits their influence in two ways. First, low-consensus samples—where false positives are more likely—have a lower probability of being selected during training. Second, DDRL assigns fixed, label-dependent advantages rather than amplifying advantages through group-relative normalization. As a result, even when false positives are sampled, their contribution to policy updates is bounded and progresses more slowly. This effect is reflected in the stabilized mean advantage dynamics shown in \Cref{fig:adv_dyn}, demonstrating that DDRL effectively suppresses the impact of false positive signals during optimization.

\section{The Probability Analysis of the Answer Frequencies and Correctness}
\label{sec:prob}

\paragraph{Probabilistic Interpretation.}
Let $x \in [0,1]$ denote the sampling frequency (self-consistency) of an answer under repeated sampling, and let $C$ and $\bar C$ denote the events that the answer is correct or incorrect, respectively. The solid curves in \Cref{fig:dist} correspond to the estimated conditional densities
\begin{equation}
p_c(x) \triangleq p(x \mid C), \qquad
p_i(x) \triangleq p(x \mid \bar C),
\end{equation}
while the dashed curve represents the posterior probability $p(C \mid x)$.

By Bayes' rule, the posterior probability of an answer being correct given its sampling frequency is given by
\begin{equation}
\begin{aligned}
    p(C \mid x)
&=
\frac{p(x \mid C)\,p(C)}
     {p(x \mid C)\,p(C) + p(x \mid \bar C)\,p(\bar C)} \\
&=
\frac{\pi_c\, p_c(x)}{\pi_c\, p_c(x) + \pi_i\, p_i(x)},
\end{aligned}
\end{equation}
where $\pi_c = p(C)$ and $\pi_i = p(\bar C)=1-\pi_c$ denote the prior probabilities.

Rewriting the posterior in log-odds form yields
\begin{equation}
\log \frac{p(C \mid x)}{1 - p(C \mid x)}
=
\log \frac{\pi_c}{\pi_i}
+
\log \frac{p_c(x)}{p_i(x)},
\end{equation}
which shows that the shape of the dashed curve is entirely determined by the likelihood ratio $p_c(x) / p_i(x)$.

As shown in \Cref{fig:dist}, incorrect answers dominate the low-frequency regime where $p_i(x) \gg p_c(x)$, resulting in $p(C \mid x) \approx 0$. Conversely, correct answers increasingly dominate at high frequencies where $p_c(x) \gg p_i(x)$, and the posterior saturates near one. Crucially, in the intermediate frequency interval---highlighted as the \emph{ambiguous region}---the two conditional densities substantially overlap, yielding a likelihood ratio close to one. In this regime, the posterior probability lies near the decision boundary and becomes highly sensitive to small variations in sampling frequency.

This behavior can be further characterized by the derivative
\begin{equation}
\begin{aligned}
& \frac{d}{dx} p(C \mid x) =\\
&p(C \mid x)\bigl(1 - p(C \mid x)\bigr)
\frac{d}{dx} \log \frac{p_c(x)}{p_i(x)},
\end{aligned}
\end{equation}
which explains why the posterior changes most rapidly in the ambiguous region while flattening at both extremes. Overall, this figure demonstrates that sampling frequency provides reliable confidence signals only in low- and high-frequency regimes, whereas medium-frequency answers inherently induce uncertainty and constitute a major source of noisy supervision in test-time reinforcement learning.

\begin{algorithm*}[t]
\caption{Debiased and Denoised Test-Time Reinforcement Learning (DDRL)}
\label{alg:ddrl}
\begin{algorithmic}[1]
\Require
Test queries $\mathcal{Q}$, pretrained policy $\pi_{\theta_0}$,
number of rollouts $N$, selected samples per query $K$,
RL epochs $E_{\mathrm{RL}}$, off-policy rollout size $M$,
SFT epochs $E_{\mathrm{SFT}}$
\Ensure
Adapted policy $\pi_{\theta}$

\State Initialize policy $\pi_{\theta} \leftarrow \pi_{\theta_0}$

\Comment{\textbf{On-policy Test-Time RL}}
\For{$e = 1$ \textbf{to} $E_{\mathrm{RL}}$}
    \For{each query $q \in \mathcal{Q}$}
        \State Sample $N$ rollouts $\{y_i\}_{i=1}^N$, where $y_i \sim \pi_{\theta}(\cdot \mid q)$
        \State Obtain pseudo-label $y^*$ via majority voting
        \State Compute frequency $c(y_i)$ for each rollout

        \Comment{\textbf{Balanced Confidence-Aware Sampling}}
        \State $K^+ \leftarrow \min\!\left(c(y^*), \left\lfloor K/2 \right\rfloor\right)$
        \State $K^- \leftarrow K - K^+$
        \State Select $K^+$ positive samples with label $y^*$
        \State Select $K^-$ negative samples with lowest frequencies

        \Comment{\textbf{Debiased Advantage Estimation}}
        \For{each selected rollout $y$}
            \State $A(y) \leftarrow \mathbb{I}(y = y^*) - \mathbb{I}(y \neq y^*)$
        \EndFor

        \State Update $\pi_{\theta}$ using policy gradient with advantages $\{A(y)\}$
    \EndFor
\EndFor

\Comment{\textbf{Consensus-Based Off-Policy Refinement}}
\State Initialize off-policy dataset $\mathcal{D}_{\mathrm{op}} \leftarrow \emptyset$
\For{each query $q \in \mathcal{Q}$}
    \State Sample $M$ rollouts $\{y_j\}_{j=1}^M$ from $\pi_{\theta}$
    \State Obtain consensus label $y^*(q)$ via majority voting
    \State Add $(q, y_j)$ to $\mathcal{D}_{\mathrm{op}}$ if $y_j = y^*(q)$
\EndFor

\For{$e = 1$ \textbf{to} $E_{\mathrm{SFT}}$}
    \State Fine-tune $\pi_{\theta}$ on $\mathcal{D}_{\mathrm{op}}$ with supervised learning
\EndFor

\State \Return $\pi_{\theta}$
\end{algorithmic}
\end{algorithm*}

\subsection{Additional Results}
\begin{figure*}[htbp]
    \centering
    \begin{subfigure}[t]{0.4\linewidth}
        \centering
        \includegraphics[width=\linewidth]{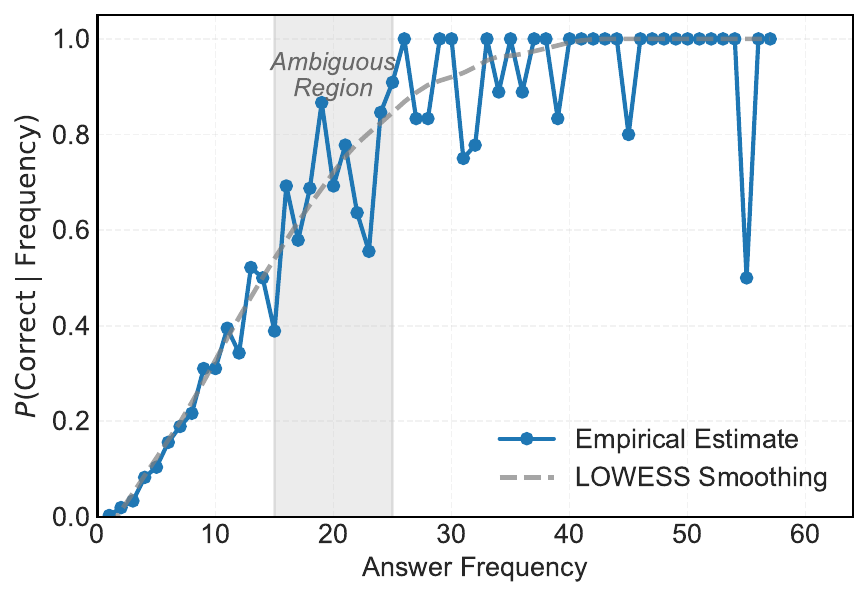}
        \caption{Correctness vs. frequency on AIME (Qwen2.5-Math-1.5B).}
    \end{subfigure}
    \hspace{25pt}
    \begin{subfigure}[t]{0.4\linewidth}
        \centering
        \includegraphics[width=\linewidth]{figs/cvpr_style_gt_rate_vs_count_smoothed_qwen2.5-math-1.5B-AIME}
        \caption{Correctness vs. frequency on AIME (Qwen2.5-3B).}
    \end{subfigure}
    \caption{Answer frequency as an imperfect proxy for reliability on AIME (1985-2024). 
}
    \label{fig:freq_AIME}
\end{figure*}
We extended the analysis to a larger AIME collection (AIME 1983–2024, 933 problems) in \Cref{fig:freq_AIME}. The same frequency–correctness pattern emerges: answers in the medium-frequency regime exhibit substantially higher uncertainty, while low-(<=5) and high-(>=45) frequency regions remain relatively reliable. This confirms that the key assumption behind BCS is not specific to MATH-500 but generalizes to other datasets.

\begin{table}[htbp]
\centering
\resizebox{\linewidth}{!}{
\begin{tabular}{lcccc}
\toprule
Method & AIME 2024 & AMC & MATH-500 & \textbf{Avg} \\
\midrule
Qwen2.5-Math-7B & 12.9 & 35.6 & 46.7 & 31.7 \\
+TTRL            & 40.2 & 68.1 & 83.4 & 63.9 \\
+DDRL            & \textbf{40.3} & \textbf{69.0} & \textbf{86.7} & \textbf{65.3} \\
\bottomrule
\end{tabular}
}
\caption{Additional Results on Qwen2.5-Math-7B.}
\label{tab:qwenmath-7b}
\end{table}
We evaluate a larger variant within the same family, Qwen2.5-Math-7B, to better isolate scaling effects in \Cref{tab:qwenmath-7b}. DDRL continues to show consistent improvements over TTRL at this larger scale

\begin{table}[htbp]
\centering
\resizebox{\linewidth}{!}{
\begin{tabular}{lccc}
\toprule
Refinement Size (compared to original setting) & AIME 2024 & AMC & MATH-500 \\
\midrule
1x (original setting) & 25.0 & 52.9 & 80.7 \\
2x                  & 25.2 & 53.0 & 80.7 \\
4x                  & 25.3 & 53.0 & 80.8 \\
\bottomrule
\end{tabular}
}
\caption{Sensitivity analysis of refinement size on mathematical reasoning benchmarks.}
\label{tab:rs}
\end{table}
To investigate the impact of the refinement size on model performance, we conduct a sensitivity analysis by scaling the original setting (1x) to 2x and 4x \Cref{tab:rs}. As shown in the table, our method demonstrates remarkable robustness to this hyperparameter. Scaling up the refinement size yields highly marginal improvements across all benchmarks. Specifically, the accuracy on AIME 2024 only increases slightly from 25.0 to 25.3, while performance on AMC and MATH-500 saturates almost immediately, showing negligible gains ($\le 0.1$). These results indicate that the model's capabilities are already fully elicited at the 1x scale. Given the linear increase in computational overhead associated with larger refinement sizes, the extremely low marginal benefit justifies our choice of 1x as the default setting. It achieves an optimal trade-off, maintaining strong reasoning performance while minimizing inference costs.